%% file: main.tex
\pdfoutput=1

\documentclass[11pt]{article}

\usepackage[final]{acl}

\usepackage{times}
\usepackage{latexsym}

\usepackage[T1]{fontenc}

\usepackage[utf8]{inputenc}

\usepackage{microtype}

\usepackage{inconsolata}

\usepackage{graphicx}

\usepackage{util}

\usepackage{float}
\usepackage{graphicx}
\usepackage{tabularx}
\usepackage{booktabs}
\usepackage{multirow}
\usepackage{adjustbox}
\usepackage{makecell}

\usepackage{xspace}

\usepackage{subcaption} 

\usepackage[ruled,vlined,linesnumbered]{algorithm2e}
\usepackage{amsmath}
\usepackage{amssymb}          

\usepackage{listings}
\newcommand{\model}{\textsc{FlowAgent}\xspace}       

\newcommand{\ofollow}{{compliance}\xspace}
\newcommand{\oflex}{{flexibility}\xspace}
\newcommand{\dinhouse}{{In-house}\xspace}
\newcommand{\codef}[1]{\texttt{#1}\xspace}
\newcommand{\metricf}[1]{\textit{#1}\xspace}

\usepackage{pifont}
\newcommand{\cmark}{\textcolor{green}{\ding{52}}}
\newcommand{\xmark}{\textcolor{red}{\ding{56}}}

\newcommand{\quotes}[1]{``#1''}

\newenvironment{envnew}{}{}

\title{\model: Achieving Compliance and Flexibility for Workflow Agents}

\author{
  \textbf{Yuchen Shi\textsuperscript{1}},
  \textbf{Siqi Cai\textsuperscript{1}},
  \textbf{Zihan Xu\textsuperscript{1}},
  \textbf{Yulei Qin\textsuperscript{1}},
  \textbf{Gang Li\textsuperscript{1}}, \\
  \textbf{Hang Shao\textsuperscript{1}}, 
  \textbf{Jiawei Chen\textsuperscript{2}},
  \textbf{Deqing Yang\textsuperscript{2}},
  \textbf{Ke Li\textsuperscript{1}},
  \textbf{Xing Sun\textsuperscript{1}} \\
  \\
  \textsuperscript{1}Tencent YouTu Lab,
  \textsuperscript{2}Fudan University \\
  \small{\textbf{Correspondence:} \href{mailto:easonsshi@tencent.com}{easonsshi@tencent.com}}
}

\begin{document}
\maketitle


\begin{abstract}
The integration of workflows with large language models (LLMs) enables LLM-based agents to execute predefined procedures, enhancing automation in real-world applications. Traditional \textit{rule-based} methods tend to limit the inherent \textit{flexibility} of LLMs, as their predefined execution paths restrict the models' action space, particularly when the unexpected, out-of-workflow (OOW) queries are encountered. Conversely, \textit{prompt-based} methods allow LLMs to fully control the flow, which can lead to diminished enforcement of procedural \textit{compliance}. 
To address these challenges, we introduce \model, a novel agent framework designed to maintain both compliance and flexibility. We propose the Procedure Description Language (PDL), which combines the adaptability of natural language with the precision of code to formulate workflows. Building on PDL, we develop a comprehensive framework that empowers LLMs to manage OOW queries effectively, while keeping the execution path under the supervision of a set of controllers. Additionally, we present a new evaluation methodology to rigorously assess an LLM agent's ability to handle OOW scenarios, going beyond routine flow compliance tested in existing benchmarks. Experiments on three datasets demonstrate that \model not only adheres to workflows but also effectively manages OOW queries, highlighting its dual strengths in compliance and flexibility. The code is available at 
\url{https://github.com/Lightblues/FlowAgent}.
\end{abstract}

\input{1_intro}
\input{5_related}

\input{2_prelim}
\input{3_model}
\input{4_data}

\input{4_exp}

\section{Conclusion}
In this paper, we reviewed existing LLM-based workflow methods and compared their strengths and weaknesses in terms of compliance and flexibility. Aiming to enhance the compliance capability of LLMs without significantly compromising their flexibility, we proposed the PDL syntax to express workflows and used the \model framework to control agent behavior. For evaluating compliance and flexibility capabilities, we constructed datasets based on existing data and designed specific evaluation methods. Experiments on three datasets demonstrated that \model not only possesses strong compliance capabilities but also exhibits robust flexibility when handling out-of-workflow queries. 

\section{Limitations} \label{sec:limitattions}

We acknowledges two primary limitations:

\paragraph{Workflow Generation} Our current research emphasizes enhancing LLM performance within manually constructed workflows using the PDL syntax. Consequently, the evaluation is limited to these artificially defined settings, lacking exploration of automated workflow generation \citep{WorFBench,AFlow}. Future work should investigate dynamic workflow synthesis to adapt to varying and complex user demands without manual intervention.

\paragraph{Dialogue Diversity and Evaluation} While this study evaluates agent performance in OOW scenarios using simulated user interactions, the real-world applicability relies on testing across a broader spectrum of authentic user demands.

\bibliography{main}

\input{6_appendix}

\end{document}

%% file: 1_intro.tex
\section{Introduction} \label{sec:intro}

\figHeader
\figSessions

With the enhanced understanding and reasoning capabilities of large language models (LLMs), pre-trained LLMs are increasingly being utilized in dialogue systems \citep{SPACE-3,TOD-Adapters}. Compared with traditional chatbots, LLMs can interact more flexibly with users to address diverse needs, leveraging the vast amount of commonsense knowledge stored in their parameters \citep{LLM-dialogue-survey}. However, in real-world applications, we often expect chatbots to follow specific rules and procedures to perform certain tasks (e.g., guiding users to make an appointment for appropriate hospitals, departments, and doctors \citep{STAR,SPACE-3}). The procedures that must be followed through dialogues are known as \emph{workflows}. LLMs, acting as \emph{workflow agents}, assist users via conversations and invoke relevant tools to fulfill requests \citep{FlowBench}.

Existing research can be broadly classified into two categories: rule-based and prompt-based methods.
\textbf{Rule-based} methods \citep{coze,dify,flowise} control the conversation between the agent and the user through deterministic programs, \emph{modeling the progress of dialogue as state transitions within a graph composed of nodes representing different dialogue states}, as shown in the upper part of Fig.~\ref{fig:header}(a). 
In this approach, the LLM functions as a node within the graph and cannot control the entire conversation flow. As a result, this method provides high compliance but often at the expense of the LLM's inherent \textbf{flexibility}. As illustrated in the lower part of Fig.~\ref{fig:header}(a), introducing a new flexible feature within this system (e.g., allowing users to pause an appointment booking process to inquire about a condition before resuming) requires the addition of numerous transition egeds (dashed lines), significantly increasing complexity.
In contrast, \textbf{prompt-based} methods leverage LLMs to autonomously manage dialogue by \emph{representing workflows textually} (natural language, code or other structured data, Fig.~\ref{fig:header}(b)). 
While this method imparts soft control over LLM responses (workflow as part of prompt), LLMs' probabilistic nature often leads to \textbf{compliance} issues, like hallucinating incorrect information, which can have serious repercussions (e.g., notifying a user about a successful appointment booking when it hasn't occurred) \citep{Hallucination-survey}.

This brings us to the critical question of our work: \textbf{How can we enhance LLM compliance with workflow tasks without diminishing their interaction flexibility?}
This question arises from two primary challenges: 1) \emph{How should we precisely represent workflows?} 2) \emph{How can we effectively control LLM behavior?}

To address the first challenge, as shown in Fig.~\ref{fig:header}(c), we introduce \emph{Procedure Description Language (PDL)}, which merges the fluidity of natural language with the precision of coding. PDL's flexible syntax allows for comprehensive node definitions, facilitating accurate workflow representations (see Sec.~\ref{subsec:pdl-syntax}).
To tackle the second challenge, we present the \model framework, which includes a set of controllers that manage agent behavior according to PDL-defined nodes. This system allows LLMs to make autonomous yet monitored and legally constrained decisions (see Sec.~\ref{subsec:arch}).

Fig.~\ref{fig:sessions} illustrates two sessions in a hospital appointment setting. In session 1, when a user wishes to switch from Hospital A to Hospital B during the registration process, \model demonstrates \emph{flexibility} by re-invoking the \codef{check\_hospital} API as per PDL directives. Conversely, in session 2, when the user prompts, \quotes{Just tell me the result of the appointment}, the LLM might attempt to respond without executing the necessary booking API. However, the controllers in the \model framework prevent such an occurrence by ensuring that prerequisite conditions are met before informing the user of the booking result, highlighting the \emph{compliance} offered by \model. 

Our \textbf{contributions} are threefold: \\
1. We provide a systematic analysis of existing LLM-based workflow agents, focusing on compliance and flexibility. Based on this analysis, we propose the PDL syntax, combining natural language and code to flexibly describe node relationships and workflow procedures. \\
2. We introduce the \model framework, which aids in the execution of workflow agents. By crafting PDL-driven controllers, we dynamically balance compliance and flexibility. Experiments on three datasets demonstrate \model's balanced compliance and flexibility within and beyond pre-defined workflows. \\
3. We construct a comprehensive evaluation benchmark augmenting existing datasets to assess workflow agent performance in out-of-workflow (OOW) scenarios. 

%% file: 5_related.tex
\section{Related Work} 
\subsection{LLM-Driven Conversational Systems} \label{subsec:llm-tod} 
The evolution of task-oriented dialogue (TOD) systems has transitioned from modular pipelines \citep{LLM-dialogue-survey} to end-to-end LLM paradigms. While traditional systems suffered from error propagation across NLU, DST, and NLG modules \citep{SPACE-3,PPTOD}, modern approaches leverage LLMs for holistic dialogue management via workflow-guided interactions \citep{FlowBench,SystemPrompt}. 
This shift necessitates new evaluation metrics focusing on task success rates over modular accuracy \citep{DAG-test-gen}, motivating our framework's dual focus on procedural compliance and adaptive flexibility.

\subsection{Agentic Workflow Architectures} 
The progression of LLMs has led to the development of LLM-based agents across various domains \citep{GenerativeAgents,MedAgents,ChatDev}.
LLM-based agents enhance task execution through tool usage and dynamic planning \citep{ReAct,Toolformer,KnowledGPT,KnowAgent}. We distinguish two paradigms: 1) \emph{Workflow generation} creates procedures via LLM reasoning \citep{AutoFlow,CoRE,LLM+P,AutoAgents,PlanBench}, and 2) \emph{Workflow execution} operates within predefined structures \citep{FlowBench,WorFBench}. 

Our research primarily focuses on the latter paradigm, treating workflows as predefined knowledge to build robust, user-centric agents. Within this context, two main approaches are adopted to integrate structured workflows with linear-text-processing language models: 
1) \emph{Rule-based Approach}: This method involves hard-coding workflow transition rules as fixed logic, defining the current node and state transitions explicitly in the program.
2) \emph{Prompt-based Approach}: Here, workflows are represented in flexible formats such as natural language, code (or pseudocode), or flowchart syntax \citep{FlowBench,KnowAgent}.
Each method presents unique challenges: rule-based systems often lack flexibility, while prompt-based methods might deviate from intended procedures. Our solution aims to strike a balance between process control and adaptability, ensuring workflows are both structured and responsive to dynamic interactions.

%% file: 2_prelim.tex
\section{Preliminary and Background} \label{sec:prelim}
\subsection{Workflow}

A \emph{workflow} defines a structured process designed to accomplish a specific task or goal within a particular scenario. For instance, in a hospital appointment booking scenario, a typical workflow involves steps such as querying the user for their preferred hospital, department, and time, retrieving available appointment slots using relevant tools, confirming the details with the user, and completing the booking. 
Formally, we can represent a workflow as a directed acyclic graph (DAG) denoted by $\mathcal{G}(\mathcal{V}, \mathcal{E})$ \citep{WorFBench,AFlow}, where $\mathcal{V}$ represents the set of nodes, each corresponding to an atomic operation (e.g., querying the user, invoking an API, retrieving from a knowledge base), and $\mathcal{E}$ represents the directed edges that define the temporal and dependency relationships between these operations, effectively specifying the workflow's progression.

\subsection{Workflow Agent} \label{subsec:workflow-agent}
A \emph{workflow agent} is designed to assist users in completing tasks by interacting with them and utilizing available tools. It can be conceptualized as an agent making sequential decisions within an environment composed of the user and the available tools. This interaction can be modeled as a Markov Decision Process (MDP), which provides a valuable framework for understanding the agent's decision-making process over time. In this framework, the agent transitions through a sequence of states ($s$), takes actions ($a$) based on the current state, and receives feedback ($r$) from the environment (user responses or tool-generated outputs). This process can be represented as $\{(s_{0}, a_{0}, r_{0}), (s_{1}, a_{1}, r_{1}), \dots, (s_{t-1}, a_{t-1}, r_{t-1})\}$. Consequently, the decision-making process of the workflow agent can be expressed as:
\begin{equation}
    a_t \leftarrow \mathcal{A}(\mathcal{H}_{t-1}, \mathcal{G}),
\end{equation}
where $\mathcal{H}_{t-1}$ encompasses all actions and observations up to time $t-1$, and $\mathcal{G}$ serves as the guide for the agent's actions.

Based on how the workflow is represented and integrated into the agent's decision-making process, workflow agents can be broadly classified into two categories: \emph{rule-based agents} and \emph{prompt-based agents}. Rule-based agents rely on explicitly programmed procedures to guide the workflow, while prompt-based agents utilize a single language model to autonomously manage the entire decision-making and dialogue process.

The first category, \textbf{rule-based agents}, implements the workflow procedure through explicit programming. Examples include Coze~\citep{coze}, Dify~\citep{dify}, and Flowise~\citep{flowise}. In these systems, the transitions between nodes are rigidly controlled by the program, with the LLM acting as a component within specific nodes to generate user responses, predict parameters for tool calls, or facilitate node transitions (e.g., classifying user intent).  In this paradigm, the information accessible to the agent and its action space are limited, which can be expressed as:
\begin{equation}
    a_t \leftarrow \mathcal{M}^{v}(\phi^{v}(\mathcal{H}_{t-1}), \psi^{v}(\mathcal{G})),
\end{equation}
where $v$ is the current node, $\phi^{v}(\mathcal{H}_{t-1})$ is the selected information visible to $v$, $\psi^{v}(\mathcal{G})$ is a subgraph of $\mathcal{G}$ expanded from $v$, and $\mathcal{M}^{v}$ denotes the language model associated with node $v$. 

The second category, \textbf{prompt-based agents} \citep{FlowBench,KnowAgent}, represents the workflow as text $\mathcal{W}^{(f)}$ using a specific format $f$, and a single language model $\mathcal{M}$ autonomously manages the entire decision-making and dialogue process. This process can be represented as:
\begin{equation}
    a_t \leftarrow \mathcal{M}(\mathcal{H}_{t-1}, \mathcal{G}^{(f)}),
\end{equation}
where $\mathcal{G}^{(f)}$ represents the graph structure implicitly encoded within $\mathcal{W}^{(f)}$. 

%% file: 3_model.tex
\tableDatasetStat
\tableDatasetCompare

\section{Methodology} \label{sec:method}
In this work, we introduce a novel procedural description language (PDL) designed to represent workflows, alongside \model, an execution framework that enhances the agent's behavioral control. 

\subsection{PDL Syntax} \label{subsec:pdl-syntax}
\begin{figure}[!t]
\begin{lstlisting}[language={}, basicstyle=\small\ttfamily] 
APIs:
- name: check_hospital
    pre: []
- name: check_department
    pre: ['check_hospital']
- name: query_appointment
    pre: ['check_department']
- name: register_appointment
    pre: ['query_appointment']
- name: recommend_other_hospitals
    pre: ['register_appointment']

    ANSWERs:
- name: inform_appointment_result
    pre: ['register_appointment']
...
- name: answer_out_of_workflow_questions
- name: request_information
\end{lstlisting}
\vspace{-0.2cm}
\caption{Example of Node Definations in PDL} \label{fig:node_example}
\vspace{-0.2cm}
\end{figure}

\begin{figure*}[!t]
\begin{lstlisting}[language={}, basicstyle=\small\ttfamily, numbers=left] 
while not API.check_hospital(hospital) or not API.check_department(hospital, department):
    hospital, department = ANSWER.request_information('hospital', 'department')
time = ANSWER.request_information('time')
available_list = query_appointment(hospital, department, time)
try:
    # ... collect necessory information for registration
    result = API.register_appointment(hospital, ...)
    ANSWER.inform_appointment_result(result)
except:
    # if registration fails, recommend other hospitals
    available_list = API.recommend_other_hospitals(department, time)
    # ... try to register again
\end{lstlisting}
\vspace{-0.2cm}
\caption{Example of Procedure Description in PDL} \label{fig:procedure_example}
\vspace{-0.2cm}
\end{figure*}

PDL consists of three primary components:
1) \textit{Meta Information}: Basic workflow details such as name and description.
2) \textit{Node Definitions}: Resources accessible to the agent, which include\emph{API} nodes (for external tool calls) and \emph{ANSWER} nodes (for user interaction).
2) \textit{Procedure Description}: The procedural logic of the task, expressed in a mix of natural language and pseudocode.

For illustration, in the \emph{Hospital Appointment} workflow, Fig.\ref{fig:node_example} presents a segment of the \emph{node definitions}
\footnote{For brevity, certain details have been omitted; see App.\ref{appendix:pdl_example} for the complete PDL specification.}.
Fig.\ref{fig:procedure_example} illustrates a portion of the \emph{procedure description}.  Key features of PDL include:
1) \textit{Precondition Specification}: Nodes include a \emph{preconditions} attribute, defining dependencies between nodes. For example, \codef{check\_department} requires \codef{check\_hospital} as a prerequisite, ensuring hospital selection before department inquiry.
2) \textit{Hybrid Representation}: The integration of natural language and code in the procedure description ensures a concise and yet flexible workflow representation, maintaining the clarity of NL with the accuracy of code.

\algo

\subsection{\model Architecture} \label{subsec:arch}

To enhance the compliance of workflow agents, we introduce \model, an execution framework tightly integrated with PDL. \model enforces a set of controllers that govern the agent's decision-making process, thereby promoting reliable action execution without sacrificing the LLM's autonomy.

Algorithm~\ref{algo} outlines \model's overall execution. Each round begins with a user query (line~\ref{alg:user}), which the agent interprets to produce a response or a tool call (line~\ref{alg:bot-action}), ultimately generating a user-facing response (line~\ref{alg:bot-response}).

To ensure decision-making stability, \model incorporates two categories of controllers: \emph{pre-decision} controllers ($\mathcal{C}_{\text{pre}} = \{c^{\text{pre}}_i\}_{i=1}^{C_{\text{pre}}}$) and \emph{post-decision} controllers ($\mathcal{C}_{\text{post}} = \{c^{\text{post}}_j\}_{j=1}^{C_{\text{post}}}$).
\textbf{Pre-decision controllers} proactively guide the agent's actions by evaluating the current state and providing feedback to the LLM (e.g., identifying unreachable nodes based on the dependency graph $\mathcal{G}^{(pdl)}$). This feedback, denoted by $\mathcal{R}_{\text{pre}}$, serves as a form of \emph{soft control}.
However, LLMs can still generate unstable outputs even with pre-decision guidance. Therefore, \textbf{post-decision controllers} provide \emph{hard constraints} by assessing the validity of proposed agent actions.  

We designed modular controllers to adjust the behavior of the workflow agent across multiple dimensions, such as \emph{enforcing node dependencies}, \emph{constraining API call repetition}, and \emph{limiting conversation length}. 
Below, using the workflow shown in Fig.~\ref{fig:node_example} as an example, we briefly introduce the \textbf{node dependency controller}.
It can operate in both pre- and post-decision modes. As a pre-decision controller ($c^{\text{pre}}_{\text{dep}}$), the controller analyzes the agent's current node and identifies inaccessible nodes by examining the dependency graph. For example, if the agent is at \codef{check\_hospital}, $c^{\text{pre}}_{\text{dep}}$ prevents the LLM from prematurely transitioning to \codef{query\_appointment} (soft control). As a post-decision controller ($c^{\text{post}}_{\text{dep}}$), the controller validates proposed node transitions.  For instance, if the agent attempts to transition to \codef{query\_appointment} without completing \codef{check\_department}, the controller denies the request, providing feedback to the agent.

%% file: 4_data.tex
\section{Evaluation and Data}

\tableExpSession
\tableExpSessionOOW

\subsection{Compliance Evaluation}
\label{subsec:eval_method}

We follow previous studies~\citep{FlowBench,AutoAgents} to conduct both turn-level and session-level assessments.
In \textbf{turn-level evaluation}, there is a reference session (considered as ground truth)~\citep{CGoDial}.
For each turn in the reference session, the evaluation system provides the prefix of the session $\mathcal{H}_{t-1}$ to the bot for predicting the current $\hat{a}_{t}$.
The judge compares $\hat{a}_{t}$ with $a_{t}$ to determine if the bot's response for that turn is correct, and the average result across all turns yields the \metricf{Pass Rate}. 
To assess the agent's tool usage capability, for turns involving tool callings, we evaluate the tool selection and parameter infilling performance of the agent in \metricf{Precision, Recall, and F1-score}.

For \textbf{session-level evaluation}, we simulate user interactions with the bot using an LLM, which serves to mimic real user behavior while minimizing human assessment costs. 
To ensure these simulated sessions accurately reflect real-world complexity, we define detailed user profiles comprising: (1) demographic information; (2) conversational style, capturing behavioral patterns; and (3) workflow-related user needs, detailing primary and secondary session objectives. 
An illustrative user profile is provided in App.~\ref{appendix:user_profile_example}. For each generated session, we conduct a binary assessment to verify whether the user's primary workflow objectives are achieved, yielding the \metricf{Success Rate}. Additionally, by tracking the number of sub-tasks initiated and completed, we derive the \metricf{Task Progress} metric. Sessions are evaluated end-to-end using prompts consistent with those recommended by \citet{FlowBench}. 
Furthermore, we evaluate the LLM agent’s performance in tool invocation with \metricf{Precision}, \metricf{Recall}, and \metricf{F1-score} metrics.

\subsection{Flexibility Evaluation}
Previous work \citep{GLAD,TRADE,AutoFlow} has primarily focused on evaluating whether bots can follow a specific procedure to complete a conversation, which partially emphasizes \ofollow while neglecting \oflex in handling user requests. Such incomprehensive evaluation may not reflect the capabilities of LLM agents under real-world scenarios, where an \quotes{imperfect} user might not adhere to the procedure and violates the sequential steps during multiple rounds of interactions. Consequently, to evaluate the performance of workflow agents in OOW scenarios, we have additionally developed a targeted evaluation method to assess \oflex.

Specifically, we categorize OOW scenarios into three types: 
(1) \textit{intent switching}, where the user suddenly changes the original intent requests or requirements, including modification of API slots/parameters and demand for cancellations; 
(2) \textit{procedure jumping}, where the user does not follow the established workflow sequence to provide information and express confirmation, including skipping steps or jumping back; 
and (3) \textit{irrelevant answering}, where the user deliberately avoids direct reply to questions raised by the agent, such as answers with topic shifts and rhetorical questions;

Based on these classifications, \oflex can be evaluated by examining the agent's performance in OOW scenarios using the metrics introduced in Sec.~\ref{subsec:eval_method}. At the turn-level, we insert OOW user interventions to assess the agent's immediate adaptive responses in these specific interactions.  At the session-level, we assess the agent's overall performance in sessions that include OOW queries to measure its long-term \oflex. 

\subsection{Data} \label{subsec:data}

We constructed three test datasets based on existing datasets and business-related data: SGD \citep{SGD}, STAR \citep{STAR}, and \dinhouse. The data construction process is detailed in App.~\ref{subsec:data_construction}. Statistics for these datasets are shown in Tab.~\ref{tab:dataset-stat}, and differences from datasets used in other studies are highlighted in Tab.~\ref{tab:dataset-compare}.

Specifically, our datasets include: (1) four types of workflows (see App.~\ref{appendix:data_examples}); (2) user profiles required for session-level evaluation (see App.~\ref{appendix:user_profile_example}); and (3) conversations needed for turn-level evaluation (see App.~\ref{appendix:conversation_example_star}).

%% file: 4_exp.tex
\figRadar 
\tableAblation

\section{Experiments} \label{sec:experiments}

We raise the following research questions: \\
\textbf{Q1}:
Compared with other models, does our proposed \model show improvements in \ofollow and \oflex? \\
\textbf{Q2}:
In which way the proposed controllers exert constraints on the model to facilitate workflows with both \ofollow and \oflex? \\

\subsection{Experimental Setup} \label{subsec:experimental_setup}

\paragraph{Baselines}
We selected ReAct \citep{ReAct} as a baseline method for comparison, which makes decisions in each round by utilizing a combination of \textit{thought} and \textit{action}, and treats the feedback from environment an \textit{observation}. It belongs to the category of prompt-based methods introduced in Sec.~\ref{subsec:workflow-agent}. For representing the workflow, we chose three formats: natural language (NL), code, and FlowChart, denoted as $\text{ReAct}_{\text{NL}}$, $\text{ReAct}_{\text{code}}$, and $\text{ReAct}_{\text{FC}}$, respectively. To ensure a fair comparison, we reused the prompts from FlowBench \citep{FlowBench} in our experiments.

\paragraph{Implementation}
In session-level evaluation, \codef{GPT-4o-mini} is used for user simulation. For the bot, we initially tested two representative model series, the GPT series \citep{GPT-4} and the Qwen series \citep{Qwen2}.
Preliminary studies revealed that small models are not competent for complex workflow tasks.
Therefore, in the present study, we choose \codef{GPT-4o} and \codef{Qwen2-72B} for demonstrations.
During the evaluation process, we used \codef{GPT-4-Turbo} for judgment.
More implementation details can be seen in App.~\ref{appendix:implementation_details}.

\subsection{Session-level Experimental Results} \label{subsec:session_level_results}

\paragraph{A1.1: \model outperforms the other three baselines in terms of task compliance.}
We first compare the session-level performance of different methods in Tab.~\ref{tab:session}.
The results indicate that \model outperforms the other three baselines in terms of task completion metrics \metricf{Success Rate}, \metricf{Task Progress}, and tool usage metrics like \metricf{Tool F1}.

\paragraph{A1.2: \model exhibits robustness towards OOW interventions with higher flexibility.}
Tab.~\ref{tab:session-oow} presents the performance of different methods under OOW scenarios.
A general performance decline is observed across all models on the three datasets.
However, \model exhibits only a slight decline, achieving the best results across all datasets.
Fig.~\ref{fig:radar}(a) visualizes the \metricf{Task Progress} metric under different settings, highlighting \model's advantage in OOW scenarios, demonstrating strong \oflex.

\subsection{Turn-level Experimental Results} \label{subsec:turn_level_results}

\paragraph{A1.3: \model maintains the superior \ofollow and \oflex across datasets in turn-level evaluation.}
We present the turn-level experimental results of Qwen2-72B in Tab.~\ref{tab:turn}. 
The results show that the \model framework achieves the best performance in both IW and OOW settings. 
What's more, Fig.~\ref{fig:radar}(b) compares the \metricf{Success Rate} across different models and settings.

\subsection{Ablation Studies}

\paragraph{A2: Controllers play an indispensable role in enforcing steady progress of workflows with OOW interventions.}
We conducted ablation experiments on \model in OOW settings, with the results shown in Tab.~\ref{tab:ab}. 
In the table, \quotes{-post} indicates the removal of the post-decision controllers $\mathcal{C}_{\text{post}}$ from the complete model, while \quotes{-post-pre} further removes the pre-decision controllers $\mathcal{C}_{\text{pre}}$. 
According to the experimental results, it is evident that removing either controller negatively impacts model performance, validating that controllers in \model enhance the model's \ofollow.

%% file: 6_appendix.tex
\clearpage
\appendix
\section*{Appendices}

\begin{envnew}
\section{Dataset Examples} \label{appendix:data_examples}

\subsection{PDL Example} \label{appendix:pdl_example}
Below is a PDL example in a real-world scenario. For formats of natural language, code and flowchat, see \citet{FlowBench}.

\begin{lstlisting}[language={}, caption=Example of PDL]
Name: 114 Hospital Appointment
Desc: Provides appointment services, allowing users to query and recommend hospitals and departments in Beijing.
Detailed_desc: Queries the availability of appointment slots based on the user's specified hospital, department, and time, and attempts to register; if no slots are available at the specified hospital, it will try to register at other hospitals.

APIs:
  - name: check_hospital
    request: [hospital_name]
    response: [hospital_exists]
    precondition: []
  - name: check_department
    request: [department_name, hospital_name]
    response: [department_exists]
    precondition: [check_hospital]
  - name: query_appointment
    request: [hospital_name, department_name, appointment_time]
    response: [available_slots, available_list, specialist_count, general_count]
    precondition: [check_hospital, check_department]
  - name: recommend_other_hospitals
    desc: Searches for available slots at other hospitals for the specified department and time.
    request: [department_name, appointment_time]
    response: [available_slots, available_list]
    precondition: [check_department]
  - name: register_hospital
    request: [id_number, appointment_type, hospital_name, department_name, appointment_time]
    response: [appointment_status]
    precondition: [query_appointment]
  - name: register_other_hospital
    request: [id_number, hospital_name, doctor_name]
    response: [appointment_status]
    precondition: [recommend_other_hospitals]

ANSWERs:
  - name: hospital_not_found
    desc: Sorry, we currently cannot provide appointment services for this hospital. Please contact the hospital directly or consider other hospitals.
  - name: department_not_found
    desc: $hospital_name does not have the department you are looking for. I will transfer you to a customer service representative for further assistance. Please wait.
  - name: no_available_slots
    desc: We apologize, but there are no available slots for the department you want to register at any hospital on our platform. Please follow the WeChat public account "Beijing 114 Appointment appointment" to register as per your needs. Thank you for calling, and have a nice day.
  - name: appointment_successful
    desc: Your appointment at $hospital_name $department_name for $appointment_time has been successful. A confirmation message will be sent to your phone number shortly. Is there anything else I can help you with?
  - name: appointment_failed
    desc: We apologize, but there are no available $appointment_type slots at $hospital_name $department_name for $appointment_time. Please follow the WeChat public account "Beijing 114 Appointment appointment" to register as per your needs. Thank you for calling, and have a nice day.
  - name: other_hospital_appointment_successful
    desc: Your appointment at $recommend_other_hospitals-hospital_name with $recommend_other_hospitals-doctor_name for $appointment_time has been successful. A confirmation message will be sent to your phone number shortly. Is there anything else I can help you with?
  - name: other_hospital_appointment_failed
    desc: We apologize, but the ID information is incorrect, and we cannot proceed with the appointment. Please follow the WeChat public account "Beijing 114 Appointment appointment" to register as per your needs. Thank you for calling, and have a nice day.
  - name: answer_out_of_workflow_questions
  - name: request_information

Procedure: |
  [hospital_exists] = API.check_hospital([hospital_name])
  if hospital_exists == false:
    ANSWER.hospital_not_found()
  elif hospital_exists == true:
    [department_exists] = API.check_department([department_name, hospital_name])
    if department_exists == false:
      ANSWER.department_not_found()
    elif department_exists == true:
      [available_slots, available_list, specialist_count, general_count] = API.query_appointment([hospital_name, department_name, appointment_time])
      if available_slots > 0:
        [appointment_status] = API.register_hospital([id_number, appointment_type, hospital_name, department_name, appointment_time])
        if appointment_status == "1":
          ANSWER.appointment_successful()
        elif appointment_status == "0":
          ANSWER.appointment_failed()
      elif available_slots == 0:
        [available_slots, available_list] = API.recommend_other_hospitals([department_name, appointment_time])
        if available_slots > 0:
          if appointment_willingness == "true":
            [appointment_status] = API.register_other_hospital([id_number, hospital_name, doctor_name])
            if appointment_status == "1":
              ANSWER.other_hospital_appointment_successful()
            elif appointment_status == "0":
              ANSWER.pther_hospital_appointment_failed()
        elif available_slots == 0:
          ANSWER.no_available_slots()
\end{lstlisting}

\subsection{User Profile Example} \label{appendix:user_profile_example}
Below is an example of a used user profile. The \quotes{User Details} contains some randomly generated attributes; \quotes{Dialogue Style} specifies the user's conversational style; \quotes{User Needs} describes the user's requirements related to a specific workflow; \quotes{Interactive Pattern} further details the possible dialogue process for the user within that workflow.

\begin{lstlisting}[language={}, caption=Example of user profile]
**Persona**:  
A 25-year-old bartender with three years of experience in the hospitality industry. He is known for his honesty, often giving customers sincere advice on their drink choices.

**User Details**:  
- Name: Michael James Carter  
- Sex: Male  
- Age: 25  
- Phone Number: 13812345678  
- ID Number: 110105199801012345  

**User Needs**:  
- Michael needs to query available appointment slots for specific hospitals and departments in Beijing.  
- He may need to verify the existence of certain hospitals and departments.  
- He wants to make an appointment for a medical consultation at a preferred hospital and department.  
- If the preferred hospital or department is not available, he may need recommendations for alternative hospitals and departments.  
- Michael may also need to know the success or failure status of his appointment registration.  

**Dialogue Style**:  
- Michael's dialogue style is likely to be straightforward and sincere, reflecting his honesty in his profession as a bartender.  
- He may prefer clear and concise information without unnecessary jargon.  
- His tone is likely to be polite and respectful but also direct, as he is used to providing sincere advice to customers.  
- He may appreciate a friendly and helpful attitude from the assistant.  

**Interactive Pattern**:  
- Michael might start by specifying the hospital and department he is interested in.  
- He is likely to ask for available appointment slots for a specific time.  
- If the hospital or department does not exist, he will appreciate being notified promptly and clearly.  
- If there are no available slots at his preferred hospital, he may ask for recommendations for other hospitals.  
- He will likely ask for the success status of his appointment registration and may need guidance on the next steps if the registration fails.  
- Michael may interact in a sequential manner, addressing one query at a time, and expecting clear and immediate responses from the assistant.
\end{lstlisting}

\section{Conversation Examples} \label{appendix:conversation_examples}

\subsection{Conversation Example on STAR Dataset} \label{appendix:conversation_example_star}
Below is a conversation generated on the STAR dataset (for turn-level evaluation). 
Note that in line 16, the user simulated by the LLM adopted an \quotes{intent-switching/detail-switching} OOW and modified the appointment time.

\begin{lstlisting}[language={}, caption=Example conversation on STAR, numbers=left]
...
BOT: Great, Alex. What's the name of the apartment you're interested in?
USER: Maple Apartments.
BOT: When would you like to view the Maple Apartments?
USER: I'd like to do it on Friday.
BOT: What time on Friday would you prefer for the viewing?
USER: How about 3 PM?
BOT: Have you paid the application fee?
USER: Yes, I have.
BOT: Would you like to add any custom message for the viewing?
USER: No, that's fine.
BOT: <Call API> book_apartment_viewing({'RenterName': 'Alex', 'Name': 'Maple Apartments', 'Day': 'Friday', 'StartTimeHour': '15', 'ApplicationFeePaid': 'Yes', 'Message': '', 'RequestType': 'CheckAvailability'})
SYSTEM: {'Status': 'Available'}
BOT: The viewing is available. Would you like to proceed with booking?
USER: Actually, I might need to change the time. Can we do 4 PM instead?
    (OOW type) intent_switching/detail-switching
BOT: <Call API> book_apartment_viewing({'RenterName': 'Alex', 'Name': 'Maple Apartments', 'Day': 'Friday', 'StartTimeHour': '16', 'ApplicationFeePaid': 'Yes', 'Message': '', 'RequestType': 'CheckAvailability'})
SYSTEM: {'Status': 'Available'}
BOT: The new time at 4 PM is also available. Shall I book it for you?
USER: Yes, please book it.
\end{lstlisting}

\section{Experimental Details} \label{appendix:exp_details}
\subsection{Implementation Details} \label{appendix:implementation_details}

For the GPT series, we specifically used the models \codef{gpt-4o-2024-05-13}, \codef{gpt-4o-mini-2024-07-18}, and \codef{gpt-4-turbo-2024-04-09}. To ensure stable output results, we set the temperature to 0.2. For the Qwen2 series models, we utilized the vllm framework for inference, also setting the temperature to 0.2.

\subsection{Prompts} \label{appendix:prompts}

\paragraph{Prompts for User Simulation}
Below are the prompts we used with LLM to simulate user behavior. The \quotes{User Profile} refers to Sec.~\ref{appendix:user_profile_example}. 

Note that for OOW simulation, we add an \quotes{additional constraints} field in the user profile to describe the user's current OOW intention. For example, \quotes{In this round, you can ask a question unrelated to the current topic} will be injected in the prompt if an \quotes{irrelevant answering} OOW intent is randomly chosen in the 5th round of dialogue. We dynamically adjust OOW queries during the conversation by controlling the injected instructions.

\begin{lstlisting}[language={}, caption=Prompt for user simulation]
You are a real-life user that interact with an assistant of {{ assistant_description }} to achieve your specific objectives. 

## User Profile
```
{{ user_profile }}
```

## History conversation
```
{{ history_conversation }}
```

## Specific requirements
1. Role Awareness: Remember you are playing the user role and speak in the first person.
2. Goal-Oriented: Keep the conversation focused on achieving your needs.
3. Style: Keep your response concise and real-life.
4. Engagement: Maintain an engaging and curious tone to facilitate effective dialogue.
5. Your output format should be:
```
Response: xxx (the response content)
```
6. Stop: End the conversation when the task is completed or when it becomes repetitive and no longer meaningful to continue. Set your response as "[END]" to stop the conversation.
\end{lstlisting}

\paragraph{Inference Prompt for \model}
Below is the inference prompt for our \model.
\begin{lstlisting}[language={}, caption=Prompt for \model]
You are a bot designed to assist the user for a specific task described by the Procedure Description Language (PDL). Your goal is to engage in a friendly conversation with the user while helping them complete the task.

### Constraints
1. **Step Identification**: Throughout the conversation, you should determine the user's current step, (whether it is in the PDL or just general questions), and dynamically follow PDL:
    - If the user's query aligns with the PDL logic, proceed to the next step.
    - If the user ask irrelevant questions, generate a response that maintains a fluent and logical conversation.
2. **PDL Components**: The PDL includes several components:
    - meta information: `name, desc, desc_detail` are meta information about the PDL.
    - slots: `slots`s define the information you may need to collect from user, or the values returned by the API.
    - reference answer: `answers` define the responses you should response to the user.
    - procedure: the final `procedure` string is a Pythonic language that defines the core logic of the procedure. 
3. Notes:
    - You have to collect enough parameter values from the user before calling the apis. 

### PDL
```PDL
{{ PDL }}
```

### Available APIs
{{ api_infos }}

### History Conversation
{{ conversation }}

### Current state
{{ current_state | trim }}

### Output Format
Your output format should be chosen from one of the two templates below. 
1. If you need to interact with the user without calling an API (inquire slot values or reply/answer):
```
Thought: xxx (description of your thought process ) 
Response: xxx (the content you need to inquire or reply)
```
2. If you need to call an API: 
```
Thought: xxx (description of your thought process ) 
Action: xxx (the function name to be called, do not prefix "API_".)
Action Input: xxx (the parameters for the function, must be in strictly valid JSON format)
```
\end{lstlisting}

\paragraph{Inference Prompt for ReAct}
For the baseline ReAct, we directly borrowed the prompt used in FlowBench \citep{FlowBench}.
\begin{lstlisting}[language={}, caption=Prompt for ReAct]
You are a helpful assistant for the task of {{task_description}}.

### Specific requirements
1. You need to act as an assistant and engage in a conversation with the user, following the business process and API information.
2. You have been provided with the flowchart information for different scenarios under a specific role.
3. You can only answer questions within the scope of the given several workflow processes. If the user asks a question beyond these scopes, please apologize and explain to the user in the response part.
4. When asking for API input parameters, ensure that the provided parameter values comply with the specified format regarding both the correctness of the format and the completeness of the content. Do not assign values arbitrarily. In instances where the parameters do not meet the format requirements, notify users to make the adjustments until the requirements are satisfied.
5. When the user has multiple requests at the same time, please select one appropriate request for processing first and inform the user that other requests will be resolved subsequently. If there is unfinished business in the previous conversation, continue to provide the necessary help and guidance to assist them in completing the business process. When multiple APIs need to be called, do so in separate rounds, with a maximum of one API call output per round. When the user indicates that the business is finished or says goodbye, respond politely and end the conversation. 

### Workflow information
```
{{workflow}}
```

### Tool information
{{toolbox}}

### Current time
{{current_time}}

### History conversation
{{history_conversation}}

### Output format
Your output format should be chosen from one of the two templates below:
1. If you need to interact with the user:
```
Thought: xxx (description of your thought process ) 
Response: xxx (the content you need to inquire or reply)
```
2. If you need to call an API (only one API call per time): 
```
Thought: xxx (description of your thought process ) 
Action: xxx (the function name to be called, do not prefix "functions.")
Action Input: xxx (the parameters for the function, must be in strictly valid JSON format)
```
\end{lstlisting}

\paragraph{Evaluation Prompts}
During the evaluation process, to ensure fairness in the results, we basically reused the prompts from FlowBench. However, for the final statistics, we only used binary results to mitigate the bias issue of the judge model (see the discussion in Sec.~\ref{subsec:eval_method}). 
Below are the prompts we used for turn-level evaluation.
\begin{lstlisting}[language={}, caption=Prompt for turn-level evaluation]
Please serve as an impartial judge to evaluate the response quality of the assistant. Your evaluation should be based on the following criteria:
(1) Correctness: Does the reply remain consistent with the workflow knowledge without any contradictions?
(2) Helpfulness: Has the user's request been reasonably understood and addressed, fulfilling the user 's needs within the provided workflow scope?
(3) Humanness: Is the response coherent, clear, complete, and does it include human acknowledgment?
Please compare the provided response with the reference response and evaluate it based on the mentioned dimensions. Then, aggregate these assessments to assign an overall score. 
A perfect score is 10 points, with 9-10 points indicating high quality, nearly identical to the reference answer; 7-8 points indicating quality close to the reference answer; 6-7 points being of moderate quality; 4-5 points indicating a lower quality response; and 2-3 points for a response with significant errors.
Finally, output a binary result to determine if the predicted and reference responses are consistent (Yes or No).

Here is the knowledge related to the workflow: 
```
{{ workflow_info }}
```

Here is the previous conversation:
```
{{ session }}
```

Here is the true value response from the reference: 
{{ reference_input }}

Here is the generated response from the assistant: 
{{ predicted_input }}

Please reply with the scores and consistency judgment in the following format:
```
Correctness Score: xxx
Helpfulness Score: xxx
Humanness Score: xxx
Consistency: Yes/No
```
\end{lstlisting}

\section{Additional Method Details} \label{appendix:method_details}

\subsection{Data Construction} \label{subsec:data_construction}

Based on existing datasets, we performed data transformation and construction to evaluate agent performance across the \ofollow and \oflex dimensions. Our data construction process consists of three stages: \textit{workflow collection, workflow representation}, and \textit{dialogue construction}.

\paragraph{Workflow Collection}
Our dataset comprises two existing datasets: SGD \citep{SGD} and STAR \citep{STAR}, as well as our own constructed dataset, \dinhouse. The SGD dataset includes 26 task flows across 16 domains, while the STAR dataset covers 24 task flows across 13 domains. The \dinhouse dataset, constructed manually based on real-world scenarios in business,
contains 6 workflows and 16 tools across 6 domains.

\paragraph{Workflow Representation}
To compare the performance of our PDL syntax with other workflow formats, we converted each workflow 
under investigation
into four formats: natural language, code, flowchart, and PDL.
Referring to \citet{FlowBench}, we first converted the workflows from the original datasets into natural language.
Then, we used a LLM to respectively transform them into code, flowchart, and PDL formats.
The definitions of tools (a.k.a., APIs) follows the OpenAI function calling formats.\footnote{\url{https://platform.openai.com/docs/guides/function-calling}}
The entire workflow format conversion process was completed using GPT-4-Turbo.

\paragraph{Dialogue Construction}
For {turn-level evaluation}, we constructed diverse user intentions from tasks, using GPT-4o to directly construct reference sessions. We then parsed and annotated tool calls at the turn level. Regarding the construction of OOW scenarios, we strategically insert OOW queries into the reference session and record the OOW information.

For {session-level evaluation}, we selected user personas from \citet{persona-hub} that exhibits real-world diversity in response style and format.
We incorporated them into workflows to construct task-related user profiles.
We employed three LLMs to respectively simulate the roles of user, agent, and system with the given user profiles, workflow descriptions, and tool definitions.
We collected these simulated dialogues to form the session-level evaluation dataset.
As for the OOW scenarios, we have simulated users generating OOW queries with a certain probability, prompting the agent to respond to these queries and continue the conversation.
The example of generated conversation is shown in App.~\ref{appendix:conversation_example_star}

\section{Additional Experimental Results} \label{appendix:experimental_results}

\subsection{Turn-level Evaluation Results} \label{appendix:turn_level_results}
The table below presents the turn-level experimental results of Qwen2-72B. It's important to note that because Out-of-Workflow (OOW) turns typically involve fewer complex conditional judgments or API calls, the turn-level \metricf{Success Rate} for OOW turns can sometimes be higher than for In-Workflow (IW) turns. Additionally, since the turn-level evaluation for the OOW portion involves fewer API calls, directly calculating this metric may introduce significant variance. Therefore, we have left it blank in the table.
\tableExpTurn

\end{envnew}